\documentclass[10pt,twocolumn,letterpaper]{article}

\usepackage[pagenumbers]{cvpr} 
\usepackage{graphicx}
\usepackage{amsmath}
\usepackage{amssymb}
\usepackage{booktabs}
\usepackage{multicol,multirow}
\usepackage{makecell}
\usepackage[table]{xcolor}
\usepackage{xurl}
\usepackage{pifont}
\usepackage{caption} 
\usepackage{multicol} 
\usepackage{tabularx}
\usepackage{listings} 
\usepackage{arydshln} 

\makeatletter
\@namedef{ver@everyshi.sty}{}
\makeatother
\usepackage{tikz}
\usepackage{bbding} 
\usepackage[weather]{ifsym}
\usepackage{fontawesome,eucal} 

\usepackage[pagebackref,breaklinks,colorlinks]{hyperref}
\usepackage{colortbl}
\definecolor{mygray}{rgb}{0.89, 0.93, 0.85}
\definecolor{whitesmoke}{rgb}{0.96, 0.96, 0.96}
\definecolor{timberwolf}{rgb}{0.86, 0.84, 0.82}

\usepackage[capitalize]{cleveref}
\crefname{section}{Sec.}{Secs.}
\Crefname{section}{Section}{Sections}
\Crefname{table}{Table}{Tables}
\crefname{table}{Tab.}{Tabs.}

\usepackage{xspace}

\newcommand{\ours}{MobileVLM V2\xspace}
\newcommand{\ourllm}{MobileLLaMA\xspace}
\newcommand{\mobilevlm}{MobileVLM\xspace}

\newcommand{\lammacppn}{\texttt{llama.cpp}}

\def\cvprPaperID{*****} 
\def\confName{CVPR}
\def\confYear{2022}

\title{MobileVLM V2: Faster and Stronger Baseline for Vision Language Model}

\author{Xiangxiang Chu\textsuperscript{1}\thanks{Equal contribution,  sorted solely by the alphabetical order of surnames}, Limeng Qiao\textsuperscript{1*}, Xinyu Zhang\textsuperscript{1*}, Shuang Xu\textsuperscript{1}, Fei Wei\textsuperscript{1}, Yang Yang\textsuperscript{1,3}, \\
Xiaofei Sun\textsuperscript{1}, Yiming Hu,\textsuperscript{1} Xinyang Lin\textsuperscript{1}, Bo Zhang\textsuperscript{1}, Chunhua Shen\textsuperscript{2}\\[.2cm]
\textsuperscript{1}Meituan Inc. ~~~~
\textsuperscript{2}Zhejiang University, China ~~~~
\textsuperscript{3}Dalian University of Technology, China
}

\begin{document}
\maketitle
\begin{abstract}
We introduce MobileVLM V2, a family of significantly improved vision language models 
upon \mobilevlm, which proves that a delicate orchestration of novel architectural design, an improved training scheme tailored for mobile VLMs, and rich high-quality dataset curation can substantially benefit VLMs' performance. Specifically, MobileVLM V2 1.7B 
achieves 
better or on-par performance on standard VLM benchmarks compared with much larger VLMs at the 3B scale. 
Notably, 
our 3B model outperforms a large variety of VLMs at the 7B+ scale. 
Our models will be released at \url{https://github.com/Meituan-AutoML/MobileVLM}.
\end{abstract}

\section{Introduction}\label{sec:intro}

To date, vision language models (VLMs)~\cite{radford2021learning,liu2023llava,Qwen-VL,zhu2023minigpt} have been a crucial research focus of the community of artificial intelligence~\cite{zhang2024mm}. Integrating large language models (LLM) with multi-modality features is verified to have unprecedented versatile capability, 
demonstrated by 
proprietary models like Gemini~\cite{fu2023gemini}, and GPT-4V~\cite{gpt4v} which have exhibited stunning performance on various downstream tasks~\cite{akter2023depth,fu2023gemini}. However, 
challenges remain to enable 
an 
vision language model as a 
capable 
intelligence assistant to be 
deployed to 
real scenarios 
such as mobile devices, 
self-driving cars, and embodied AI systems, \emph{etc}. 


Most recently, MobileVLM~\cite{chu2023mobilevlm} comes first in exploring the capacity of VLMs at the mobile scale with innovative hardware-oriented architectures. MoE-LLaVA~\cite{lin2024moellava} manages to adapt the mixture-of-experts method ~\cite{jacobs1991adaptive,eigen2013learning} 
for 
VLMs which outrageously pushed the limits of smaller models to yield surpassing performance compared with much larger ones. 
As the latest survey~\cite{zhang2024mm} indicates, the progress of VLMs exhibits a trend of extended modalities, refined training pipelines, efficient architectures, and higher-quality training datasets.

In this paper, we establish faster and stronger baselines built upon MobileVLM \cite{chu2023mobilevlm}. Our key improvements are made on mainly three aspects, namely, 
exploiting contributive training data on small VLMs, exploring effective training strategies, and renovating a high-performance lightweight projector. 
Specifically, 
we utilize $1.2$ million high-quality image-text pairs conducted by ShareGPT4V \cite{chen2023sharegpt4v} to effectively align vision-language features, and incorporate more academic tasks to increase data diversity and instruction-following capacity, such as ScienceQA \cite{lu2022learn}, TextVQA~\cite{singh2019towards}, SBU~\cite{Ordonez_2011_im2text}, \textit{etc.}. 
As for the training paradigm, we conduct thorough training of all parameters of projector and language model during both the pretraining and instruction tuning stages, which proves advantageous in harnessing the full potential of superior-quality data.
Additionally, we introduce a more streamlined yet potent projection mechanism that bridges vision and language models. By improving the representation of image tokens with enhanced positional information, we can significantly compress the number of image tokens without much performance degradation.

\begin{figure}[t]
  \centering
   \includegraphics[width=.97\columnwidth]{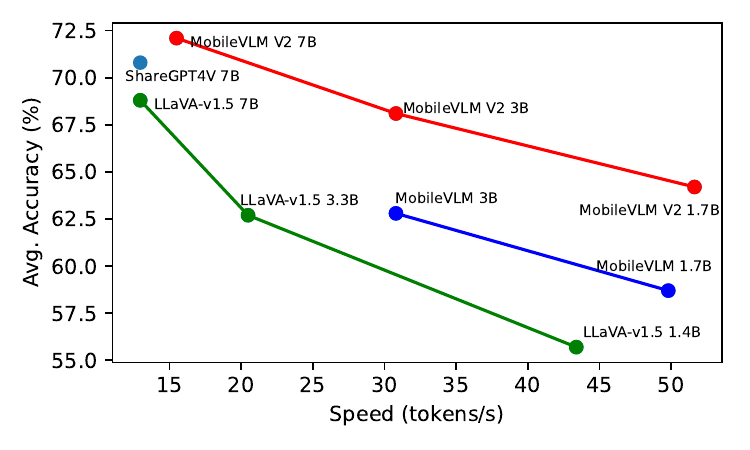}
   \caption{Comparison of SOTA VLMs in terms of average performance across several standard benchmarks and speed (tested on an NVIDIA Jeston Orin with \lammacppn). \ours achieves new state-of-the-art results with much faster inference speed.}
   \label{fig:latency-comparison-mobile}
\end{figure}

Our main contributions are summarized as follows:
\begin{enumerate}
    \item We explore and evaluate the performance of increasing training data for small vision language models, which significantly bridges the gap between small VLMs 
    such as 
    MobileVLM \cite{chu2023mobilevlm} and large ones.
    \item 
    We dive into better training strategies for mobile scenarios and design a novel training scheme on how to fully exploit the potential of more high-quality multimodal data.
    We propose a very lightweight projector to significantly reduce the visual tokens 
    with slight performance drops.
    \item Our method achieves a new state-of-the-art tradeoff between performance and inference speed across several vision language benchmarks. By scaling our model to 7B parameters, our method outperforms previous SOTA models with clear 
    margins.
\end{enumerate}

\section{Related Work} 
\label{sec:related}

\par \noindent \textbf{Large Language Models.}
Over the last few years, the emergence of LLMs has led to significant advancement in natural language processing, including GPT-3\cite{brown2020language}, PaLM\cite{chowdhery2022palm}, OPT\cite{zhang2022opt}, and BLOOM\cite{scao2022bloom}. A representative work of democratic LLMs is LLaMA, which further enhances the enthusiasm for research on improving LLM effects. Following the representative works InstructGPT\cite{ouyang2022training} and ChatGPT\cite{ChatGPT}, Alpaca and Vicuna fintuned LLaMA with human-guided annotations to enhance the language interaction abilities. Recently, many 
organizations 
have proposed their large language models \cite{gpt4, yang2023baichuan, bai2023qwen, 2023internlm, du2022glm}, which have been evaluated on multiple benchmarks and achieved excellent performance. 
Due to limited computational resources of edge devices 
such as 
mobile phones, autonomous driving cars, and embodied AI systems, several lite-LLMs~\cite{GALACTICA,biderman2023pythia,li2023textbooks,tinyllama,chu2023mobilevlm} have attracted considerable attention. It is worth noting that TinyLLaMA \cite{tinyllama} conducts great open-source work on their 1.1B base and chat models. MobileLLaMA \cite{chu2023mobilevlm} scales down the architecture of LLaMA and releases 1B and 3B models trained from scratch with public datasets. Meantime, model compression 
techniques including 
quantization and pruning on large language models is 
actively studied too 
\cite{frantar2023sparsegpt,frantar2022gptq,xiao2023smoothquant}. Lately, the precision of LLMs is reduced to W4A8 \cite{li2023speed} or even to binary \cite{li2023norm} for faster 
inference with minimal accuracy drop. 

\par \noindent \textbf{Multimodal Large Language Models.}
Several works \cite{tan2019lxmert, radford2021learning, li2021align} have proposed a series of multimodal model architectures, which are primarily composed of the vision encoder and language model. \cite{cho2021unifying, wang2022image} explore visual language reasoning tasks under a unified transformer architecture. The paradigm of the vision-language model demonstrates better generalizability than traditional vision models designed for specific tasks.
With the rapid development of LLMs, many works \cite{Qwen-VL, liu2023llava, zhu2023minigpt, chen2023sharegpt4v} have concentrated on infusing visual knowledge into LLMs. FROMAGe\cite{koh2023fromge} and LLaVA\cite{liu2023llava} directly feed visual tokens to LLMs, thereby enabling the LLM to comprehend semantic information of images and correctly respond to queries for visual contents. InternLM-XComposer2 \cite{internlmxcomposer2} proposes a Partial LoRA (PLoRA) approach to preserve the integrity of pre-trained language knowledge, which strikes a balance between precise vision understanding and text composition with literary talent. Its highly competitive performance shows remarkable proficiency in the realm of multimodal understanding. To address the urgent demand for deployment of MLLMs on edge devices, several works~\cite{fu2023gemini, chu2023mobilevlm, lin2024moellava, zhu2024llavaphi, wei2023vary} have been proposed. Gemini~\cite{fu2023gemini}, a leader in this field, has released lightweight vision-language models Gemini Nano with 1.8B/3.25B parameters, which are specifically designed for smartphones. However, their models and data are not open-sourced. MobileVLM~\cite{chu2023mobilevlm} may be the first approach to offer open-source 1B/3B vision-language models under resource-constrained scenarios. MoE-LLaVA~\cite{lin2024moellava} proposes an MoE-based sparse model architecture to enhance visual understanding capabilities. LLaVA-Phi \cite{zhu2024llavaphi} utilizes Phi--2.7B as a language foundation model to achieve superior performance in vision language reasoning tasks. Vary \cite{wei2023vary} introduces an improved vision vocabulary and 
%
enhance the image features, leading to better 
generality. 
Notably, Vary \cite{wei2023vary}
demonstrates superior potential in fine-grained perceptual tasks.

\par \noindent \textbf{Multimodal Instruction-Tuning.}
To address the modality gap between vision and language, works such as\cite{radford2021learning} employ contrastive learning to align visual and textual feature representations. Furthermore, 
methods in \cite{lai2023lisa, pi2023detgpt} leveraged the capabilities of large language models to achieve universal detection or segmentation with strong zero-shot capabilities. Inspired by the success in natural language processing, Liu et al.\ \cite{liu2023llava} introduced visual instruction tuning aiming at creating a general-purpose multimodal model with language as task instructions. SVIT\cite{zhao2023svit} scales up visual instruction tuning by constructing a dataset of 4.2 million visual instruction tuning data, which is featured by the high quality and rich diversity generated by prompting GPT-4 with the abundant manual annotations of images. ShareGPT4V\cite{chen2023sharegpt4v} constructs 100K high-quality captions from GPT4-Vision for instruction tuning, which have been expanded to 1.2 million detailed and informative captions for pre-training, covering world knowledge, object attributes, spatial relationships, and aesthetic assessments. LVIS-instruct4v\cite{wang2023see} builds 220K visually aligned and context-aware instructions produced by prompting the powerful GPT-4V with images from LVIS.
 
\section{Method}
\label{sec:method}

\begin{figure}[t!]
  \centering
   \includegraphics[width=\linewidth]{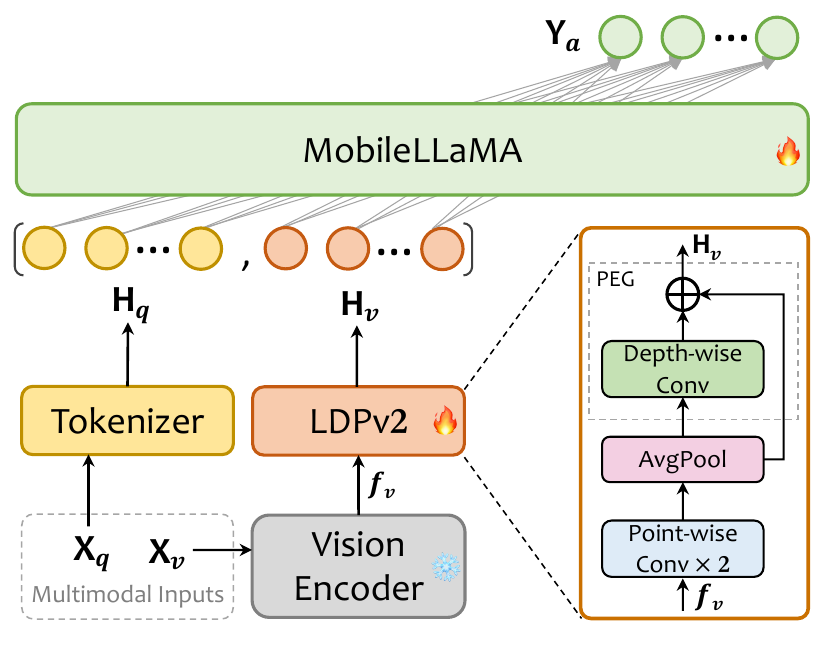}
   \caption{\textbf{\ours's architecture.} $\mathbf{X}_v$ and $\mathbf{X}_q$ indicate image and language instruction, respectively, and $\mathbf{Y}_a$ refers to the text response from the language model \ourllm. The diagram in the lower right corner is a detailed description of LDPv2, \ie, the lightweight downsample projector v2.}
   \label{fig:mobilevlmv2-arch}
\end{figure}

Our method follows a similar framework as MobileVLM \cite{chu2023mobilevlm}.
As shown in Figure~\ref{fig:mobilevlmv2-arch}, the overall architecture of \ours consists of a pre-trained vision encoder to extract image features, a pre-trained large language model \ourllm to process multi-modal tokens and generate final answers, and a mobile-friendly projector, \ie, lightweight downsample projector (denoted as LDPv2), to align image features with the language model. We introduce the details of each component in the following.

\subsection{Vision Encoder} 

Following \mobilevlm, we use CLIP ViT-L/14 \cite{radford2021learning} as the vision encoder $\mathbf{F}_{enc}$, which is contrastively pre-trained on millions of image-language pairs and proven to be effective for VLMs \cite{chu2023mobilevlm}. Particularly, images $\mathbf{X}_v \in \mathbb{R}^{H\times W\times C}$ are first resized to a resolution of 336 $\times$ 336, and flattened into patches with a stride $P$ of 14. High-level visual embeddings $f_v \in \mathbb{R}^{N_{v}\times D_{v}}$ are then extracted to represent the semantic information of images, where $N_{v}={HW/P^{2}}$ and $D_v$ denote the sequence length and the hidden size of visual embeddings. Formally we have,
\begin{equation}
{f_v}=\mathbf{F}_{enc}(\mathbf{X}_v).
\end{equation}

\subsection{Language Model}

We employ the MobileLLaMA ~\cite{chu2023mobilevlm} series as the foundational large language model (LLM). This choice has three advantages. Firstly,  MobileLLaMA is designed to facilitate off-the-shelf deployment and has demonstrated real-time speed on resource-limited devices with impressive performance. Keeping the language model unchanged helps us conduct controlled experiments to explore many other aspects such as scaling up the data corpus, improving training strategy, renovating new designs for projectors, and so on. Secondly, MobileLLaMA shares the same tokenizer with LLaMA2, which helps to perform distillation without any pain. Last but not least, it is trained on open datasets and has no risk of evaluation pollution from data leakage. This also helps to confirm whether and how this model can outperform other proprietary counterparts.  

Specifically, we adopt MobileLLaMA-1.4B-Chat and MobileLLaMA-2.7B-Chat. 
The text input $\mathbf{X}_q$ is first tokenized and processed to the text tokens $\mathbf{H}_{q}\in \mathbb{R}^{N_{t}\times D_{t}}$, where $N_{t}$ denotes the sequence length of text tokens and $D_{t}$ is the hidden size of the word embedding space. 
The text tokens $\mathbf{H}_q$ and the visual tokens $\mathbf{H}_v$ are transformed by the projector, to be concatenated as the input of the language model. The final response $\mathbf{Y}_{a}$ with a length $L$ is generated in an autoregressive manner as follows,
\begin{equation}
p({\mathbf{Y}_{a}}|{\mathbf{H}_{v}},{\mathbf{H}_{q}})=\prod_{i=1}^{L}p(y_{i}|{\mathbf{H}_{v}},{\mathbf{H}_{q}},y_{<i}).
\end{equation}

\subsection{Lightweight Downsample Projector}
\label{sec:light_weight_projector}

Inspired by the LDP design of MobileVLM\cite{chu2023mobilevlm}, we introduce a new projector to perform better vision-language feature alignment with fewer parameters. It contains three components, \textit{i.e.}, \textit{feature transformation}, \textit{token reduction}, and \textit{positional information enhancement}. 
First, we employ two point-wise convolution layers on image tokens to match the feature dimension of LLM. Then, we introduce an average pooling layer to extremely compress the number of image tokens. 
Finally, a very simple but effective module PEG \cite{chu2023conditional} with skip connection is applied to enhance positional information. 
Compared with LDP \cite{chu2023mobilevlm}, this positional part is more efficient and reduces \textbf{99.8\%} number of parameters, 
and is slightly faster in running speed.

In formulation, the lightweight downsample projector LDPv2 (connoted as $\mathbf{P}$) transforms the visual embedding $f_v\in \mathbb{R}^{N_{v}\times D_{v}}$ to the modality-aligned visual tokens $\mathbf{H}_{v}$ with positional enhancement. It is worth noting that this design is also deployment-friendly since it is composed of well-supported operators by mainstream inference frameworks. Given an average kernel $k$, the number of remaining tokens is only ${1}/{k^{2}}$ of the input features. Concretely, we formulate the proposed LDPv2 as follows,
\begin{equation}
\mathbf{H}_{v}=\mathbf{P}(f_v)= \begin{cases}
  f_{0}&=PW(GELU(PW(f_v))) \\  
  f_{1}&=AvgPool_{2\times2}(f_{0})\\
\mathbf{H}_{v}&=DW(f_{1}) + f_{1}.
\end{cases}
\end{equation}
where $PW$ and $DW$ are pointwise and depthwise convolutions  respectively, $GELU$ is a GELU~\cite{hendrycks2016gaussian} activation layer, and $AvgPool_{2\times2}$ is a $2\times2$ average pooling layer.

\begin{table}[htp]
\setlength{\tabcolsep}{2pt}
\begin{center}
\begin{small}
\begin{tabularx}{0.47\textwidth}{lXXXXXX}
\toprule
\multirow{2}{*}{Method} & \multicolumn{3}{c}{Pre-training}               & \multicolumn{3}{c}{Multi-task training}     \\
                      & $\mathcal{V}$ & $\mathcal{P}$ & $\mathcal{L}$ & $\mathcal{V}$ & $\mathcal{P}$ & $\mathcal{L}$ \\
\midrule
LLaVA-1.5-7B~\cite{liu2023improved}            &  \ding{55}    & \checkmark    &  \ding{55}      &  \ding{55}     & \checkmark      & \checkmark \\       
ShareGPT4V-7B~\cite{chen2023sharegpt4v}             &  \checkmark\llap{\tiny\ding{55}}     & \checkmark    & \checkmark     &   \ding{55}      & \checkmark      & \checkmark \\
MobileVLM~\cite{chu2023mobilevlm}              &  \ding{55}     & \checkmark    &  \ding{55}      &  \ding{55}     & \checkmark      & \checkmark \\       
\rowcolor{mygray}
\ours                 &  \ding{55}   & \checkmark    & \checkmark     & \ding{55}     & \checkmark      & \checkmark    
\\
\bottomrule
\end{tabularx}
\caption{\textbf{Comparision of training strategy.} $\mathcal{V}$, $\mathcal{P}$ and 
 $\mathcal{L}$ respectively represent the vision encoder, the projector, and the language model. \checkmark indicates that the corresponding model parameters are optimized in training, \ding{55} signifies that the parameters are frozen, and \checkmark\llap{\tiny\ding{55}} represents that a subset of model parameters is frozen.}  
\label{tab:train-strategy}
\end{small}
\end{center}
\end{table}

\subsection{Training Strategy}
Our training process is split into two stages: pre-training and multi-task training.
As illustrated in Table~\ref{tab:train-strategy}, unlike previous training paradigm of LLaVA-1.5~\cite{liu2023improved} and MobileVLM~\cite{chu2023mobilevlm}, \ours opens the projector and large language model consistently in both stages, with the visual encoder frozen.

\subsubsection{Pre-training}

Most Vision-Language Models (VLMs)~\cite{liu2023improved, cha2023honeybee} commonly freeze the visual encoder and the language model during pre-training to avoid optimization difficulties. ShareGPT-4V~\cite{chen2023sharegpt4v} partially freezes the visual encoder and makes the language model trainable.
In our case, we initialize the weights of the visual encoder and language model from CLIP ViT-L/14~\cite{radford2021learning} and MobileLLaMA~\cite{chu2023mobilevlm}, respectively.   This initialization serves as a robust foundation for the subsequent unified training process. We allow for full training of the projector and the LLM while fixing the vision encoder. Note that freezing ViT also reduces the training cost. The model's training objective is then concentrated on the prediction of the next token, utilizing an autoregressive loss function. By honing in on this specific task, the model is better equipped to learn the intricacies of language generation in the context of visual information, leading to improved performance on multimodal tasks.

As 
reported 
in Table~\ref{tab:dataset}, during the pretraining stage, our model was trained 
using 
the ShareGPT4V-PT dataset~\cite{chen2023sharegpt4v}, comprising 1.2 million image-text pairs. This dataset is instrumental in enhancing the model's image-text alignment capabilities, which is a critical aspect of multimodal representation learning.

\begin{table}[htbp]
\centering
\setlength{\tabcolsep}{1.5mm}{
\begin{tabular}{lll}
\toprule
Datasets                               & Type          & Samples          \\
\midrule
\rowcolor{whitesmoke} \multicolumn{3}{l}{\textit{\textbf{Pretraining}}}                \\
ShareGPT4V-PT~\cite{chen2023sharegpt4v}                         & Caption       & 1.2M             \\
\rowcolor{whitesmoke} \multicolumn{3}{l}{\textit{\textbf{Multi-task training}}}                \\
Visual Dialog~\cite{das2017visual}                          & Conversation        & 123K             \\
Text-VQA~\cite{singh2019towards}                            & VQA(Open)           & 35K              \\
VSR~\cite{Liu2022VisualSR}                                  & VQA(Open)           & 13K              \\
VIGC~\cite{wang2023vigc}                                    & VQA(Open)           & 37K              \\
IConQA~\cite{lu2021iconqa}                                  & VQA(MC)             & 107K             \\
SQA~\cite{lu2022learn}                                      & VQA(MC)             & 13K              \\
COCO~\cite{chen2015microsoft}                               & Caption             & 592K             \\
SBU~\cite{Ordonez_2011_im2text}                             & Caption             & 844K             \\
ShareGPT4V~\cite{chen2023sharegpt4v}                        & Mixed               & 665K             \\

\rowcolor{mygray}
\textit{\textbf{Total}}                                 &               & 3.6M             \\

\bottomrule

\end{tabular}}
\caption {\textbf{Datasets used for \ours training.} Column  {``}Type{"} is the task of the dataset, where {``}Mixed{"} indicates that the dataset contains samples belonging to various tasks. Column {``}Samples{"} is the number of image-text pairs in each dataset.}
\label{tab:dataset}
\end{table}

\subsubsection{Multi-task Training}~\label{sec:it_strategy}

After the pre-training phase of image-text alignment learning, \ours has acquired a rudimentary capability to comprehend image content. However, it lacks proficiency in utilizing visual information for analysis and dialogue in a suite of downstream tasks. Consequently, during the multi-task training phase, we introduce multiple vision-language tasks, engaging parameters in the training process to endow the model with the capacity for multi-task analysis and image-text conversing.

In the multi-task training phase, we employ a plethora of datasets featuring a variety of tasks to further develop the model's array of skills. As outlined in Table~\ref{tab:dataset}, these datasets are meticulously chosen to improve conversational abilities using the Visual Dialog dataset~\cite{das2017visual}, OCR skills through the TextVQA dataset~\cite{singh2019towards}, scene understanding capabilities via COCO Caption~\cite{chen2015microsoft} and SBU~\cite{Ordonez_2011_im2text} datasets, and location understanding by the VSR dataset~\cite{Liu2022VisualSR} \etc. Note that we have cleaned and refined the SBU dataset, where the data volume might not match exactly with the officially released one. In total, the aggregated data for this phase consists of 2.4 million samples, ensuring a comprehensive learning experience across different modalities and tasks. 
Examples regarding the dialogue formats across the various datasets are provided in the Appendix \ref{apd:dataset_sample}.

\section{Experiment}
\label{sec:exp}
\subsection{Training}
\textbf{Pre-training}
As listed in Table~\ref{tab:hyperparam}, LDPv2 is randomly initialized in this stage, and the language model and the vision encoder are initialized with the pre-trained weights from \ourllm and CLIP ViT-L/14, respectively. 
For optimization, we utilize the AdamW optimizer~\cite{loshchilov2017decoupled} with no weight decay. Maximum learning rates for LDPv2 and other components are configured as $1e^{-3}$ and $2e^{-5}$, respectively, following a cosine learning rate schedule. 
The pre-training phase involves a global batch size of 256 across 8 NVIDIA A100 GPUs for about 5 hours.

\textbf{Multi-task Training}
During this training stage, the weights of \ours are initialized from the first stage. Most training hyper-parameters are similar to the pre-training phase. Additionally, the learning rate is set to $4e^{-5}$. Training in this stage requires 8 NVIDIA A100 GPUs for around 9 hours with a global batch size of 128. Detailed training settings are shown in Table~\ref{tab:hyperparam}.

\begin{table}[htbp]
\setlength{\tabcolsep}{2pt}
\begin{center}
\begin{small}
    \begin{tabular}{lcc}
         \toprule
         Configuration            & Pre-training & Multi-task training \\
         \midrule
         Vision encoder init                 & CLIP ViT-L & CLIP ViT-L \\
         LLM init                 & MobileLLaMA & \ours PT \\
         Projector init           & Random & \ours PT  \\
         Image resolution         & $336^2$ & $336^2$ \\
         Image token num          &144  &144  \\
         Global batch size        & 256 & 128 \\
         Training steps           & 5K & 19K \\
         Optimizer                & AdamW & AdamW \\
         LR schedule   & Cosine decay & Cosine decay \\
         Projector LR  & $1e^{-3}$ & $4e^{-5}$ \\
         Base LR       & $2e^{-5}$ & $4e^{-5}$ \\
         Weight decay             & 0 & 0 \\
         Warm-up ratio            & 0.03 & 0.03  \\
         DeepSpeed Stage          & 2 & 3 \\
         \bottomrule
    \end{tabular}
    \caption{\textbf{Training hyperparameters of \ours.} Note that \ours PT indicates the parameters saved after the pre-training phase. LR denotes learning rate.
    }
    \label{tab:hyperparam}
\end{small}
\end{center}
\end{table}


\subsection{Comparisons with State-of-the-art Methods}

Following \mobilevlm,  we adopt a list of benchmarks including image question answering series GQA  \cite{hudson2019gqa}, SQA \cite{lu2022learn}, TextVQA \cite{singh2019towards}, comprehensive benchmarks MME \cite{fu2023mme}, MMBench \cite{liu2023mmbench}, and object hallucination benchmark POPE \cite{li2023evaluating}. 

\textbf{Comparison with SOTA methods.}
We evaluate the performance of \ours and show the accuracy result in Table~\ref{tab:compare-with-sotas-vlms}. Note that previous models mainly focus on accuracy improvement, regardless of the run time latency. Although our models are targeted for real applications regarding two aspects, they outperform most of the previous models with clear margins. With an advantage of 75\% faster speed, \ours 3B still outperforms a very recent work MoE-LLaVA-2.7B$\times$4 \cite{lin2024moellava} by 1.4 points on the average score. Noting MoE-LLaVA-2.7B$\times$4 \cite{lin2024moellava} already achieves comparable or better accuracy over many 7B+ VLMs. Many of them introduce large amounts of training costs. In contrast, our method achieves new state-of-the-art results while our training cost is comparable to computation-friendly LLaVA-1.5\cite{liu2023improvedllava}.

\begin{figure}[htbp]
  \centering
   \includegraphics[width=.97\columnwidth]{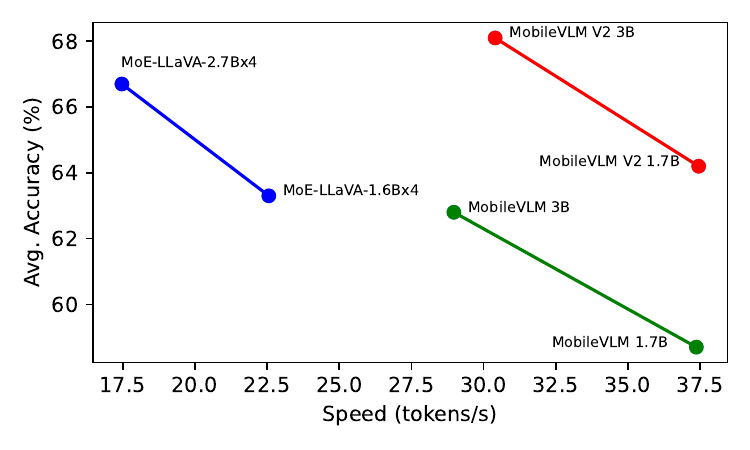}
   \caption{Speed comparison of SOTA VLMs on an NVIDIA A100 GPU. The accuracy is averaged on six VLM benchmarks (see  Table~\ref{tab:compare-with-sotas-vlms}).  (tested with a batch size of 1, generating 256 tokens).}
   \label{fig:latency-comparison}
\end{figure}

\textbf{Latency Comparison.} Since many models haven't been supported by advanced mobile inference frameworks, we compare the latency of a larger spectrum of recent models using the PyTorch framework on an NVIDIA A100 GPU, as shown in Figure~\ref{fig:latency-comparison}. MobileVLM V2 models are generally faster and stronger in terms of token generation and average scores on tested benchmarks. Especially, MoileVLM V2 1B/3B is 37.37 tokens/s and 28.97 tokens/s respectively, which are both 1.65$\times$ faster than their counterparts of MoE-LLaVA, yet with higher average performance. 


\begin{table*}[t!]
\centering
\setlength{\tabcolsep}{3pt}
\small 
\begin{tabular}{*{1}{l}*{2}{l}|*{7}{c}}
\toprule
Method & LLM & Res. & GQA & SQA$^\text{I}$ & VQA$^\text{T}$ & POPE & MME$^\text{P}$ & MMB$^\text{dev}$ & Avg.\\
\midrule
IDEFICS-80B \cite{laurenccon2023obelisc} & LLaMA-65B & 224  & 45.2 & -- & 30.9 & -- & -- & 54.5 &-- \\
BLIP-2 \cite{li2023blip2} & Vicuna-13B & 224  & 41.0 & 61.0 & 42.5 & 85.3 & 1293.8 & -- & -  \\
InstructBLIP \cite{dai2023instructblip}& Vicuna-13B & 224 & 49.5  & 63.1 & 50.7 & 78.9 & 1212.8 & -- & -  \\
Shikra \cite{chen2023shikra}& Vicuna-13B & 224  & -- & -- & -- & -- & -- & 58.8 & --  \\

Openflamingo \cite{anas2023OpenFlamingo} & MPT-7B & 336 & -- & -- & 33.6 & -- & -- & 4.6 & -- \\
Qwen-VL \cite{Qwen-VL} & Qwen-7B & 448& 59.3 & 67.1 & 63.8 & -- & 1487.6 & 38.2 & - \\

mPLUG-Owl \cite{ye2023mplug} & LLaMA-7B & 224  & -- & -- & -- & -- & 967.3 & 49.4 & -- \\
IDEFICS-9B \cite{laurenccon2023obelisc} & LLaMA-7B & 224& 38.4 & -- & 25.9 & -- & -- & 48.2 & -- \\

MiniGPT-v2 \cite{chen2023minigpt} & LLaMA-7B & 448 & 60.3 & -- & -- & -- & -- & 12.2 & --\\
MiniGPT-4 \cite{zhu2023minigpt} & Vicuna-7B & 224  & 32.2 & -- & -- & -- & 581.7 & 23.0 & -- \\
InstructBLIP \cite{dai2023instructblip} & Vicuna-7B & 224  & 49.2  & 60.5 & 50.1 & -- & -- & 36.0 & -- \\
LLaVA-1.5 \cite{liu2023improved} & Vicuna-7B & 336  & 62.0 & 66.8 & 58.2 & 85.9 & 1510.7 & 64.3 & 68.8 \\
ShareGPT4V \cite{chen2023sharegpt4v} & Vicuna-7B & 336  & 63.3 & 68.4 & 60.4 & 85.7 & 1567.4 & 68.8 & 70.8 \\

MoE-LLaVA-1.6B$\times$4 \cite{lin2024moellava} & StableLM-1.6B& 336 & 60.4& 62.6 & 47.8 &84.3 & 1300.8$^\text{*}$ & 59.4 & 63.3 \\
MoE-LLaVA-2.7B$\times$4 \cite{lin2024moellava} & Phi-2.7B& 336 &  61.1& 68.7 & 50.2 &85.0 & 1396.4$^\text{*}$ & 65.5 & 66.7 \\
\midrule
MobileVLM 1.7B \cite{chu2023mobilevlm} & MobileLLaMA 1.4B & 336& 56.1 & 57.3 & 41.5 & 84.5 & 1196.2 & 53.2 & 58.7 \\
\rowcolor{mygray} \ours{ 1.7B} & \ourllm{ 1.4B} & 336 &59.3&66.7&52.1&84.3&1302.8&57.7 & 64.2 \\
\midrule
MobileVLM 3B \cite{chu2023mobilevlm} & MobileLLaMA 2.7B & 336  & 59.0 & 61.2 & 47.5 & 84.9 & 1288.9 & 59.6 & 62.8 \\
\rowcolor{mygray} \ours{ 3B}& \ourllm{ 2.7B} &336 &61.1&70.0&57.5&84.7&1440.5&63.2& 68.1\\
\midrule
\rowcolor{mygray}\ours{ 7B}& Vicuna-7B & 336 &62.6&74.8&62.3&85.3&1560.7&69.2&72.1\\
\rowcolor{mygray}\ours { 7B} w/o AvgPool& Vicuna-7B & 336 &64.6& 74.8&66.8&86.1&1558.7&70.8& 73.5\\
\bottomrule
\end{tabular}
\caption{\textbf{Comparison with SOTA methods on six VLM benchmarks.} GQA  \cite{hudson2019gqa}; SQA$^\text{I}$: ScienceQA-IMG  \cite{lu2022learn}; VQA$^\text{T}$: TextVQA  \cite{singh2019towards}; POPE  \cite{li2023evaluating}; MME$^\text{P}$: MME Perception \cite{fu2023mme}; MMB$^\text{dev}$: MMBench-dev  \cite{liu2023mmbench}; Column \textit{Res.} is the image resolution of vision model. Column \textit{Avg.} indicates the average accuracy on six evaluation benchmarks. 
The values in the MME$^\text{P}$ column should be divided by 2000 before being included in the average accuracy calculation.
* represents that results not provided in the paper \cite{lin2024moellava} were evaluated using their latest repo (commit 5ba14e8). The value of \mobilevlm on SQA$^\text{I}$ is 
erroneous  
in \cite{chu2023mobilevlm} and 
is corrected 
in this version.
}
\label{tab:compare-with-sotas-vlms}
\end{table*}

\textbf{Comparison with MoE-LLaVA. } MoE-LLaVA utilizes several experts to achieve good performance, and each expert is a small model to improve the inference speed. Although only a proportion of parameters is activated, it still requires storing the whole parameters, which inevitably incurs IO overhead in mobile scenarios. Moreover, it's nontrivial to apply model compressing tricks \cite{li2023norm, frantar2022gptq, chu2023make, zhou2024LiDAR-PTQ} to these models. In contrast, our method can be well-supported and optimized for deployment. In a word, our inference speed advantage evaluated on the Tesla A100 GPU will be further enlarged if tested on resource-limited environments. In principle, \ours can also be combined with its MoE design. But how to combine it without sacrificing the memory and latency advantage of \ours remains as our future work.

\textbf{Comparison with \mobilevlm.}
 Table~\ref{tab:compare-with-sotas-vlms} shows that   \ours significantly improves the accuracy performance of \mobilevlm. \textit{The average accuracy is boosted by 5.3 points}. Since these two methods share the same encoders both for vision and language modalities, we attribute the improvement to the enhanced data, training strategies, and the new design of the projector. This good performance of \ours indicates that MobileLLaMA from \cite{chu2023mobilevlm} is also a good baseline for small language models because it's built on the open resource dataset Redpajama~\cite{together2023redpajama}, which is reproducible and has a low risk of evaluation data leakage. Exploring open and more powerful small language models remains as our future work.
 

\subsection{Model Scaling Analysis}

\begin{figure}[ht]
  \centering
   \includegraphics[width=.97\columnwidth]{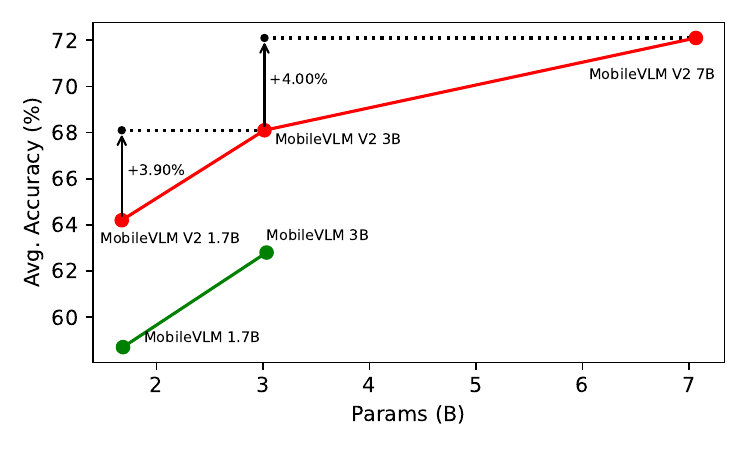}
   \caption{Average performance improvement on six VLM benchmarks when scaling \ours's models across several standard tasks (see also Table~\ref{tab:compare-with-sotas-vlms}).
   }
   \label{fig:model-scaling}
\end{figure}

\begin{figure}[ht]
  \centering
   \includegraphics[width=.97\columnwidth]{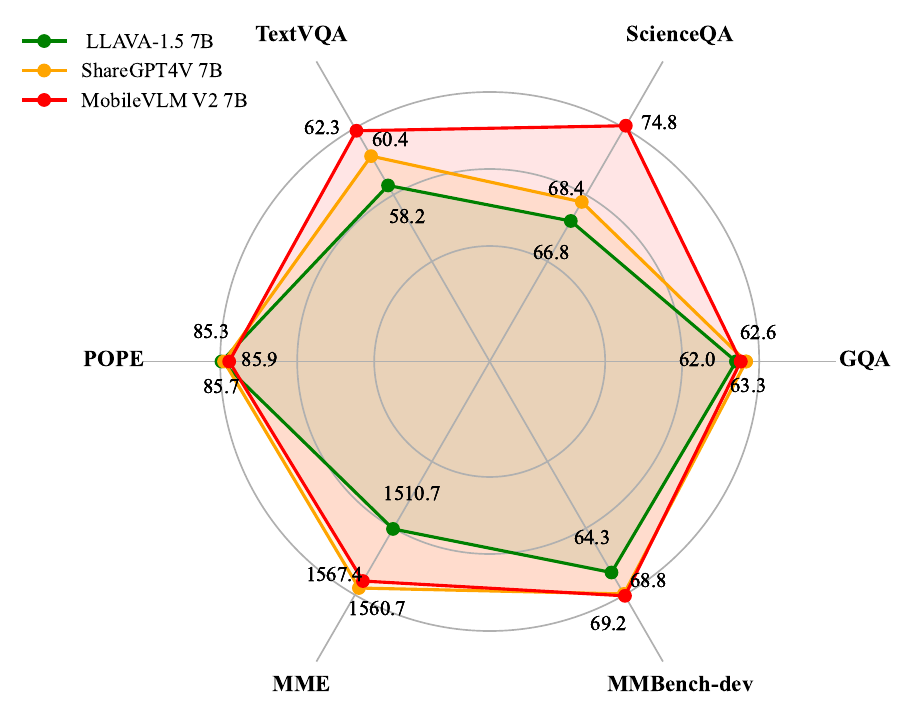}
   \caption{Radar plot of \ours's performance compared with its peers on 6 standard benchmarks.}
   \label{fig:model-radar}
\end{figure}

While our target is to design powerful multimodal models for resource-constrained scenarios, we further scale up our model to verify the upper bound of performance.  This also forms an apple-to-apple comparison with many existent VLMs. Specifically, we utilize Vicuna-7B as the LLM model and show the result in 
Figure \ref{fig:model-scaling}, where a comprehensive multimodal performance improvement emerges.
We compare our \ours 7B with mainstream large-scale VLMs like LLaVA-1.5 7B \cite{liu2023improved} and ShareGPT4V 7B \cite{chen2023sharegpt4v} in both accuracy and the inference speed, which is shown in Figure \ref{fig:model-radar} and Table \ref{tab:mobile-vlm-latency-Orin}. It can be seen that our \ours 7B not only secures a significant performance edge across multiple benchmarks but also establishes a clear lead in inference speed over large-scale representatives. While being nearly 20\% faster, \ours 7B outperforms ShareGPT4V by 1.3 points regarding the average performance.  
This further illustrates the effectiveness of our data scaling strategy, training strategy, and the novel projector design.

Given the observation that the latency gap is 
narrowed 
upon the 7B model, we remove the token reduction component, \emph{i.e.}, average pooling with a kernel of 2. Under this setting, our 7B model has the same latency speed as ShareGPT4V. We show the result in the last row of Table~\ref{tab:compare-with-sotas-vlms}. \ours 7B (w/o AvgPool) achieves an average score of 73.5, which significantly outperforms LLaVA-1.5 by 4.7 points. Compared with \ours 7B (with AvgPool), the performance gain mainly comes from the increased score on the TextVQA task, which is an OCR task. This task contains many small objects where token reduction could be harmful. We retain how to effectively make use of high-resolution input in our future work.

\subsection{Latency Measurement on Mobile Devices}
\label{sec:latency}
To maintain consistency, we measure the inference latency of  \ours on the NVIDIA AGX Jetson Orin platform with identical configurations as MobileVLM~\cite{chu2023mobilevlm}. The \lammacppn~\cite{Gerganov2019} framework is employed as the inference framework. Specifically, for the Jetson Orin platform, we develop more efficient CUDA implementations for LDPv2 (Section \ref{sec:light_weight_projector}) to fully leverage the hardware to have the best performance. Table~\ref{tab:mobile-vlm-latency-Orin} reports the comparison results of inference latency. To more objectively evaluate the inference performance, we introduce $Eval_{avg}$, which denotes the actual generation speed of output tokens, calculated by dividing the total time by the number of 256 output tokens. 

\textbf{Inference Latency on NVIDIA Jetson Orin.}
On the Jetson Orin platform, we conclude that \ours demonstrates a lower inference latency than its counterparts at the same parameter scale. As shown in Table~\ref{tab:mobile-vlm-latency-Orin}, \ours achieves the best performance 
in the inference speed, which can be attributed to a more lightweight design of the projector: we optimize the original 576 tokens of visual prompts to 144 ones, while the performance on the average accuracy remains the same or 
becomes even better. 

\begin{table}[ht]
  \centering
  \setlength{\tabcolsep}{1pt}
  \begin{tabular}{*{1}{l}|*{3}{c}}
    \toprule
    Model     &\makecell[c]{Avg.\\accuracy}  & \makecell[c]{$Eval_{avg}$\\(tokens/s)} & $Total$ (s) \\
    \midrule
    ShareGPT4V-7B & 70.8 & 13.00 & 19.69 \\
            LLaVA-1.5 7B & 68.8 & 12.96 & 19.75  \\
            \rowcolor{mygray}
            \ours 7B & \textbf{72.1} & {15.49} & {16.53}\\
            LLaVA-1.5 3.3B & 62.7 & 20.45& 12.52 \\
            MobileVLM 3B & 62.8 & 30.80 & {8.31} \\
            \rowcolor{mygray}
            \ours 3B & 68.1 & {30.80} & 8.38 \\
            LLaVA-1.5 1.4B & 55.7 & 43.39 & 5.90 \\
            MobileVLM 1.7B & 58.7 & 49.80 & {5.14} \\
            \rowcolor{mygray}
            \ours 1.7B & 64.2 & \textbf{51.63} & \textbf{4.96} \\
    \bottomrule
  \end{tabular}
  \caption{Latency comparison of mobile-scale VLMs on NVIDIA Jetson Orin. The language model of VLMs is quantized to 4-bit with \lammacppn. The average accuracy is evaluated on the same six benchmarks as Table~\ref{tab:compare-with-sotas-vlms}. }
  \label{tab:mobile-vlm-latency-Orin}
\end{table}


\begin{table}[ht]
\centering
\setlength{\tabcolsep}{1.8pt}
\begin{tabular}{*{1}{l}|*{2}{l}|*{7}{c}}
\toprule
 &D&T & GQA & SQA$^\text{I}$ & VQA$^\text{T}$ & POPE  & MME$^\text{P}$ & MMB$^\text{dev}$ & Avg.\\
\midrule
(a)  &\ding{55}&\ding{55}& 56.1 & 57.3  & 41.5 & 84.5 & 1196.2 & 53.2 &58.7\\
(b) &\checkmark &\ding{55}& 57.5 & 63.9&  49.8 &83.9 & 1157.5&  51.6  &60.8\\
\rowcolor{mygray}
(c) &\checkmark &\checkmark &58.5&  65.4 &  50.8& 83.4 &1262.6  &55.4&62.8\\

\bottomrule
\end{tabular}
\caption{Ablation results of Data Strategy and Training Strategy on \mobilevlm 1.7B. `D' and `T' represent our Data Strategy and Training strategy, respectively.} 
\label{tab:quantitive_of_data_training_strategy}
\end{table}

\begin{table*}[htp]
	\centering
	\setlength{\tabcolsep}{4pt}
    \scalebox{0.90}{
	\begin{tabular}{*{1}{l}|*{3}{c}|*{6}{c}|*{1}{c}}
		\toprule
		VL Projector Architecture Design  & w/o D\&T & Tokens & Params. & GQA & SQA$^\text{I}$ & VQA$^\text{T}$ & POPE  & MME & MMB$^\text{dev}$ & Avg.  \\
		\midrule
        $[{PW}]_{\times 2}$ &\ding{55} & 576 & 6.30M& 56.9 & 57.1 & 43.7 & 85.7 & 1137.7 & 52.8 & 58.8 \\
        $[{PW}]_{\times 2}[DW^{\kappa = 1}PW]_{\times 1}[DW^{\kappa = 2}PW]_{\times 1}$ &\ding{55} & 144  & 18.94M &56.1 & 57.3 & 41.5 & 84.5 & 1196.2 & 53.2 & 58.7 \\
        \midrule
		$[{PW}]_{\times 2}$ & \checkmark & 576 & 6.30M & 59.9 & 63.7 & 53.9 & 85.0 & 1271.3 & 56.0  & 63.7  \\		
		$[{PW}]_{\times 2}[DW^{\kappa = 1}PW]_{\times 1}[DW^{\kappa = 2}PW]_{\times 1}$ & \checkmark & 144 & 18.94M & 58.5 & 65.4 &  50.8& 83.4 &1262.6  &  55.4 & 62.8 \\
        $[{PW}]_{\times 2}[{AvgPool}^{\rho = 2}]_{\times 1}$ & \checkmark & 144 & 6.30M & 59.3 & 65.0 & 53.1& 84.0& 1292.2& 54.5 & 63.2 \\
        $[{PW}]_{\times 2}[{AvgPool}^{\rho = 2}]_{\times 1}[{LearnablePE}]_{\times 1}$ & \checkmark & 144 & 6.59M & 59.1 & 67.1 & 52.3 & 84.3 & 1286.7& 55.5 & 63.7 \\
        \rowcolor{mygray}  $[{PW}]_{\times 2}[{AvgPool}^{\rho = 2}]_{\times 1}[DW^{\kappa = 1}]_{\times 1}$ & \checkmark & 144 & 6.32M & 59.3&66.7&52.1&84.3&1302.8&57.7 & 64.2 \\
		\bottomrule
	\end{tabular}
    }
	\caption{The exploration of projector design based on \ourllm 1.4B. The $PW$ represents pointwise-$conv$ and $DW$ is depthwise-$conv$. The subscript $\times$ indicates the number of times the corresponding module is stacked repeatedly. The superscript $\kappa$ indicates the $3\times3 \ conv$ stride and $\rho$ indicates the pooling kernel size. Note that D represents our proposed Data Strategy and T for Training Strategy in \ours.
    The\colorbox{mygray}{green row} is the proposed LDPv2 in our \ours.}
	\label{tab:ablation-on-projector}
\end{table*}

\section{Ablation Study}
\subsection{Effect of Data Scaling}
\label{sec:data_effect}

In this section, we explore the effect of the data enhancement strategy. As shown in Table \ref{tab:quantitive_of_data_training_strategy}, the first row represents the baseline of \mobilevlm 1.7B. We replace the pretraining dataset with the ShareGPT4V dataset ~\cite{chen2023sharegpt4v} and expand the instruction tuning dataset to 2.4M, as described in Section \ref{sec:it_strategy}. Note that the training strategy and model architecture remain the same as the baseline. Performance gains on GQA, SQA, and TextVQA demonstrate that the model's ability in cognition and dialogue has been improved after data enhancement. However, we also observe that in the configuration of more high-quality data, maintaining the original training strategy cannot fully exploit the data benefits and leads to performance degradation on MME and MMBench.

\subsection{Effect of Training Strategy}

Based on the observations from Section \ref{sec:data_effect}, we further explore a more reasonable training strategy. Under the setting of enhanced data and small-scale VLMs, unlocking the language model in the whole training stage enables more effective learning.
Therefore, we opt for simultaneous finetuning of the projector as well as the language model.
Compared results are shown in the second and third rows in Table \ref{tab:quantitive_of_data_training_strategy}. We can see that this training setup enables a comprehensive improvement in model performance across most VLM benchmarks, with a 2-point gain on average accuracy.

\subsection{Effect of Projector}
Looking back at MobileVLM, the proposed LDPv1 reduces the number of tokens by 75\% (576 $\to$ 144) while maintaining almost equivalent performance, see the first and second rows in Table \ref{tab:ablation-on-projector}. Based on the newly proposed data configuration and training strategy, the above corresponding architectures can achieve $5.4$ and $4.5$ average improvements respectively (the third and fourth rows).
However, when we try to continue to explore better alignment methods of visual and language features, we observe that increasing the number of learnable parameters can easily lead to optimization dilemmas, resulting in a decline in the overall alignment effect.
Based on this observation, we firstly replace the $[{DW}^{\kappa}PW]$ block in LDPv1 with an embarrassingly simple parameter-free operation, \ie, $2\times2$ average pooling, and find that we could obtain an average performance improvement of 0.4 points (see row 5 in Table \ref{tab:ablation-on-projector}).
Secondly, to make the aligned visual features with stronger positional information, the simplest idea is to employ learnable positional encoding for enhancement, which brings $0.5$ performance gains in the 6\textit{th} row.
Further, inspired by the claims of PEG \cite{chu2023conditional}, whose position encoding is dynamically generated and conditioned on the local neighborhood of the input tokens, we then replace the learnable PE with a PEG layer to obtain better-enhanced features. The seventh row in Table \ref{tab:ablation-on-projector} indicates an improvement of 0.5 points.
It is worth mentioning that, a PEG layer applied in \ours 1.7B only contains $0.02M$ ($2048\times3\times3$) parameters. Compared with the $[{DW}^{\kappa}PW]$ block in MobileVLM, the number of learnable parameters has dropped by nearly $630\times$ ($12.64M \to 0.02M$), but the overall performance is improved by 1.4 points, which demonstrates the effectiveness of our design.

\section{Conclusion}
In a nutshell, we propose a series of efficient vision language models called \ours based on MobileVLM\cite{chu2023mobilevlm}. We explore data scaling schemes, improved training strategies, and efficient modality alignment design to improve the overall performance under the setting of small VLM models. With a comparable training cost as \cite{liu2023improved}, our method achieves a new state-of-the-art Pareto front in terms of accuracy and latency, targeted for real product environments. Our model outperforms many larger models with substantial inference advantages, which paves a promising way to enjoy advanced AI on resource-limited 
scenarios.

\paragraph{Acknowledgements:}   This work was in part 
supported by National Key R\&D Program of China (No.\  2022\-ZD\-0118\-700).
\label{sec:concl}

{
    \small
    \bibliographystyle{ieeenat_fullname}
    \bibliography{mobilevlm-v2-main}
}

\clearpage
\clearpage
\appendix
\section{Dialogue formats of various datasets.}
\label{apd:dataset_sample}

During the pre-training phase, we utilized the 1.2 million image-text pairs from the pre-training phase of ShareGPT4V, which primarily includes COCO~\cite{lin2014microsoft}, SAM~\cite{kirillov2023segment}, and LLaVA-1.5 pre-training data~\cite{liu2023improved}.
Subsequently, in the multi-task training phase, we collected 2.4 million high-quality instruction data. 
A detailed format of these data is provided in Table~\ref{tab:data formats1} and Table~\ref{tab:data formats2}.

\section{Examples of \ours.}
\label{apd:inference_sample}
\begin{figure*}[t!]
  \centering
   \includegraphics[width=\linewidth]{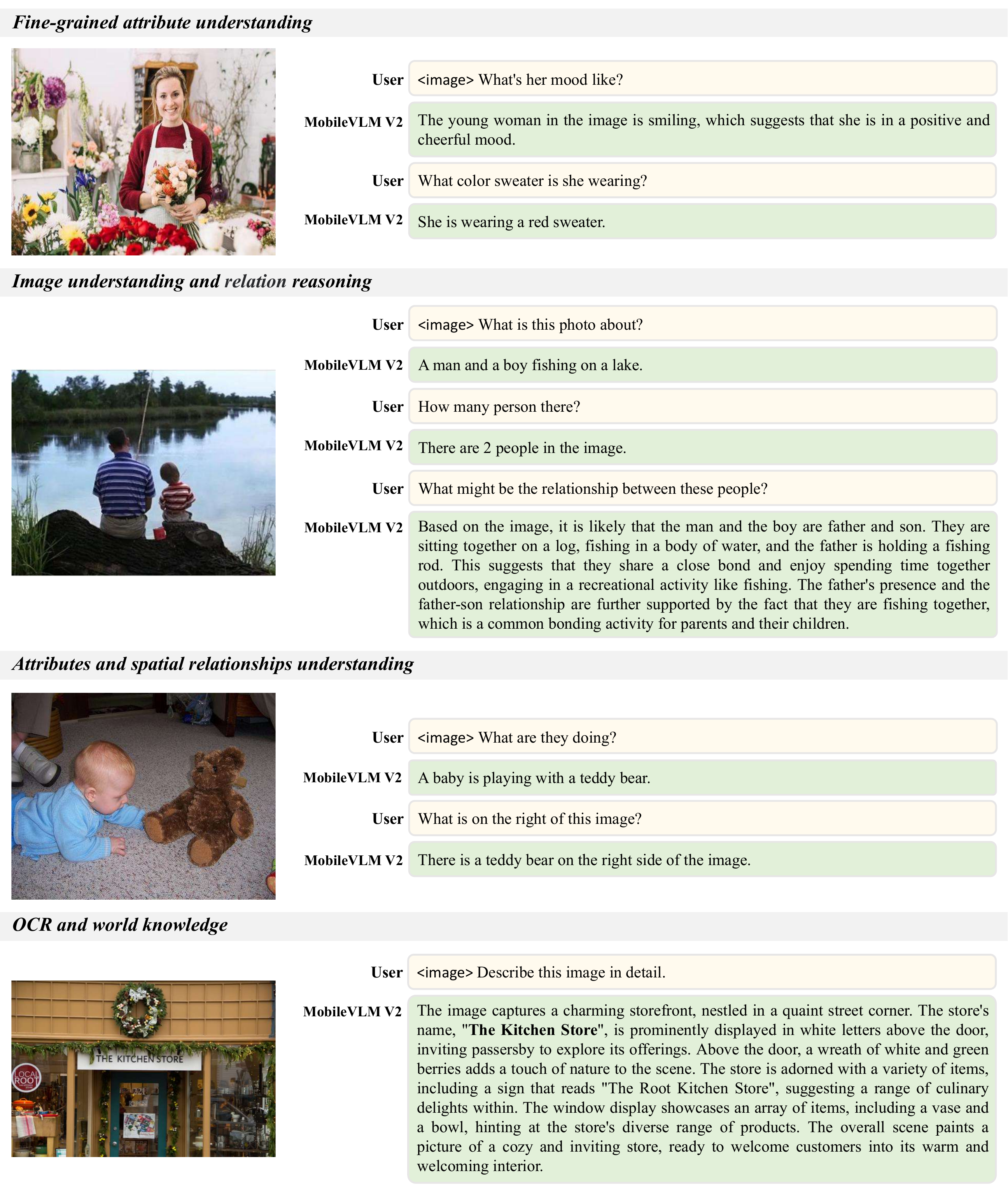}
   \caption{Examples of \ours 1.7B in various scenes. }
   \label{fig:inference_sample}
\end{figure*}

Figure~\ref{fig:inference_sample} shows qualitative results of \ours 1.7B in multimodal conversations in various scenarios including of fine-grained attribute understanding, image understanding and relation reasoning, attributes and spatial relationships understanding, OCR and world knowledge, \etc.

\clearpage 
\onecolumn 

\sloppy
\begin{table}[htbp]
\centering
\begin{tabularx}{\textwidth}{ccX}
\hline
\textbf{Dataset} & \textbf{Image} & \textbf{Dialogue Content} \\ \hline \\

\multirow{11}{*}{VIGC} & 
\multirow{11}{*}{\includegraphics[width=0.3\textwidth]{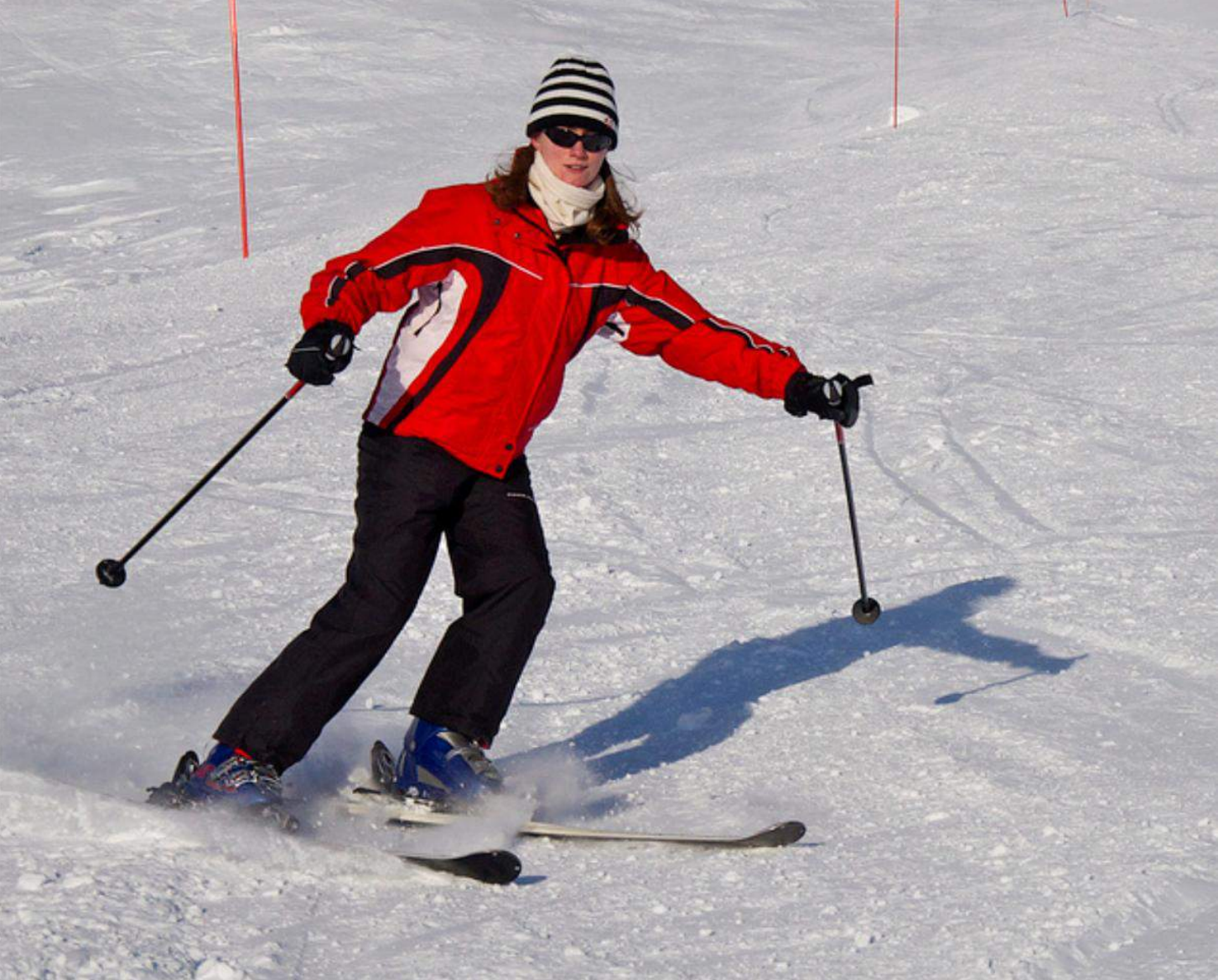}} & 
\textbf{User:} \texttt{<image>\textbackslash n}What is the person in the image doing? \\
&& \textbf{Assistant:} The person in the image is skiing down a snow-covered slope. \\ \\ \\ \\ \\ \\ \\ \\ \\ \hdashline \\

\multirow{5}{*}{IConQA} & 
\multirow{5}{*}{\includegraphics[width=0.3\textwidth]{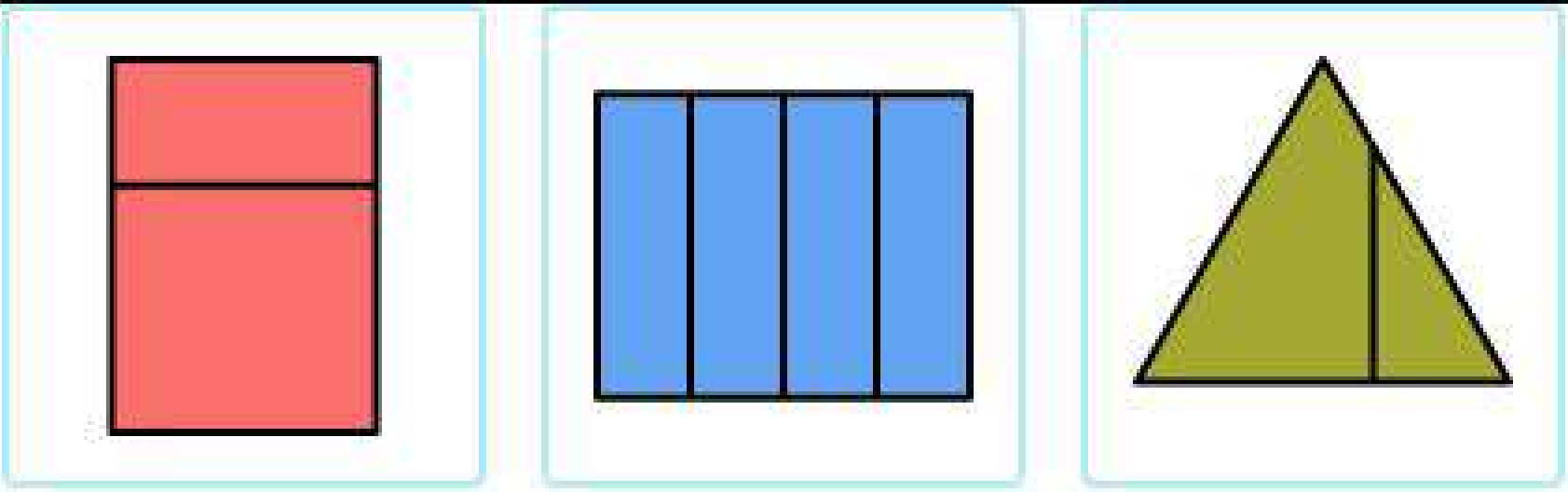}} & 
\textbf{User:} \texttt{<image>\textbackslash n}Select the picture that shows equal parts.A. First image.\texttt{\textbackslash n}B. Second image.\texttt{\textbackslash n}C. Third image.\texttt{\textbackslash n}Answer with the option's letter from the given choices directly. \\
&& \textbf{Assistant:} C. \\ \\ \hdashline \\

\multirow{10}{*}{TextVQA} & 
\multirow{10}{*}{\includegraphics[width=0.3\textwidth]{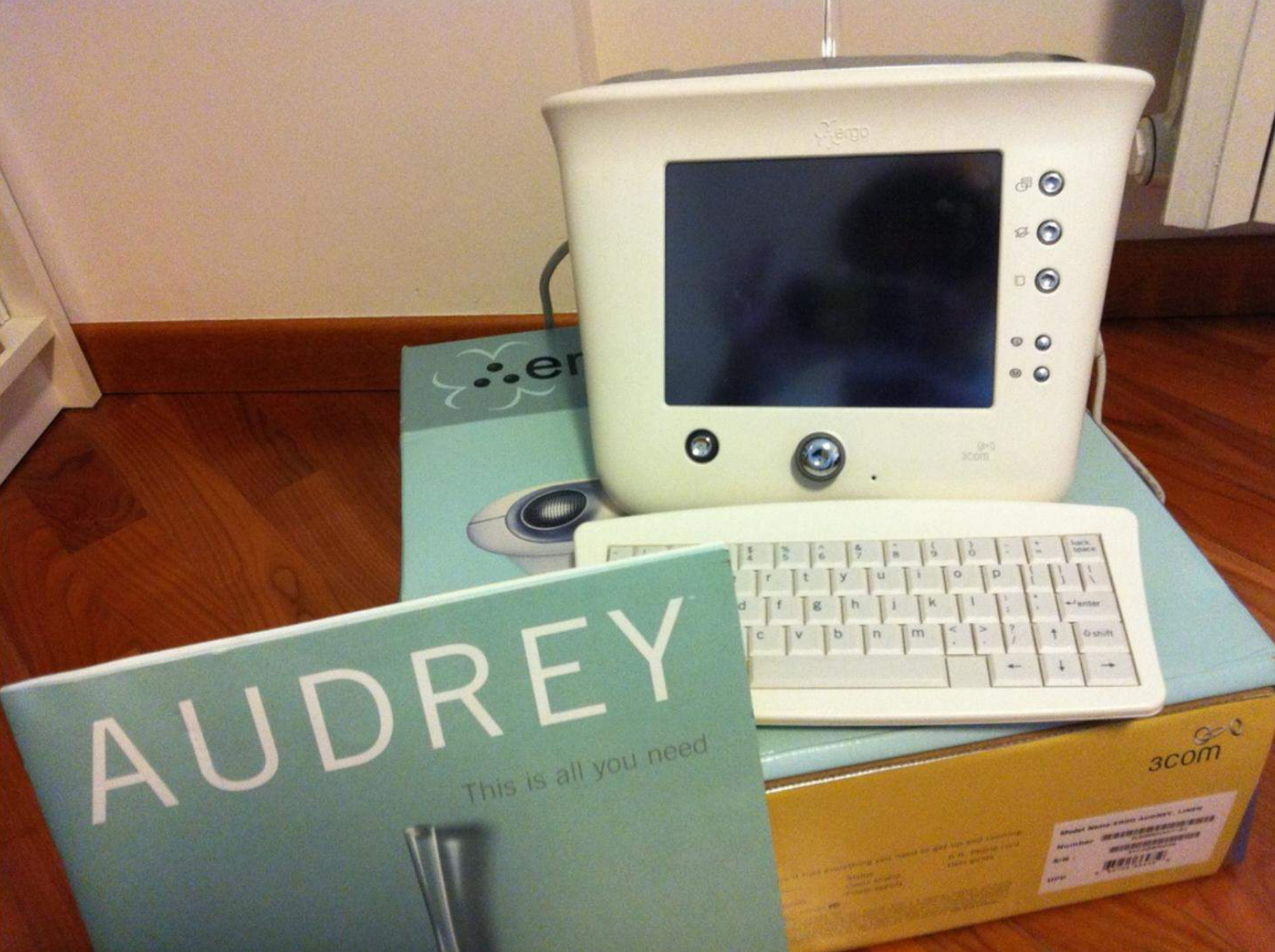}} & 
\textbf{User:}\texttt{<image>\textbackslash n}What brand name is on the box?\texttt{\textbackslash n} Reference OCR token: Cergo, AUDREY, is, all, you, need, 3com, This \texttt{\textbackslash n}Answer the question using a single word or phrase. \\
&& \textbf{Assistant:} 3com. \\ \\ \\ \\ \\ \\ \\ \hdashline \\

\multirow{19}{*}{COCO Caption} & 
\multirow{19}{*}{\includegraphics[width=0.3\textwidth]{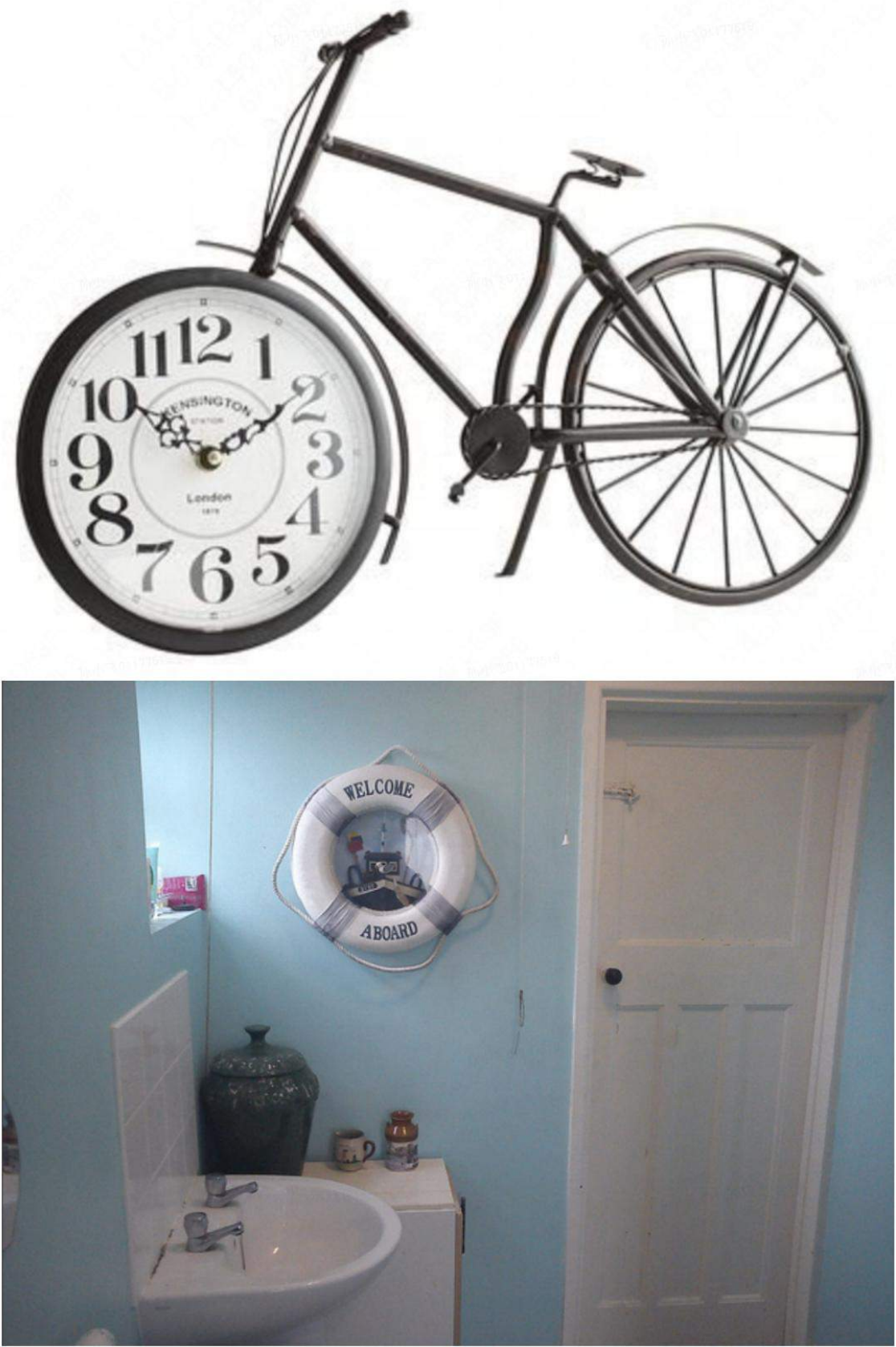}} & 
\textbf{User:} \texttt{<image>\textbackslash n}What are the key elements in this picture? \\
&& \textbf{Assistant:} A bicycle replica with a clock as the front wheel. \\ \\ \\ \\ \\ \\ \\ \\
&& \textbf{User:} \texttt{<image>\textbackslash n}What is this photo about? \\
&& \textbf{Assistant:} A room with blue walls and a white sink and door. \\ \\ \\ \\ \\ \\ \\ \\ \\ \hline
\end{tabularx}
\caption{Examples from the multi-task training data for \ours (Part 1).}
\label{tab:data formats1}
\end{table}

\clearpage 
\sloppy
\begin{table}[htbp]
\centering
\begin{tabularx}{\textwidth}{ccX}
\hline
\textbf{Dataset} & \textbf{Image} & \textbf{Dialogue Content} \\ \hline \\
\multirow{9}{*}{Visual Dialog} & 
\multirow{9}{*}{\includegraphics[width=0.3\textwidth]{figures/visualdialog.pdf}} & 
\textbf{User:} \texttt{<image>\textbackslash n}How many bikes there? \\
&& \textbf{Assistant:} 3. \\
&& \textbf{User:} What color are bikes? \\
&& \textbf{Assistant:} I see green red and white. \\
&& \textbf{User:} Are they parked on stock parking? \\
&& \textbf{Assistant:} NO. \\ 
&& \textbf{User:} What color of their helmets? \\
&& \textbf{Assistant:} green and white and other is black. \\ \\ \hdashline \\

\multirow{11}{*}{SBU} & 
\multirow{11}{*}{\includegraphics[width=0.3\textwidth]{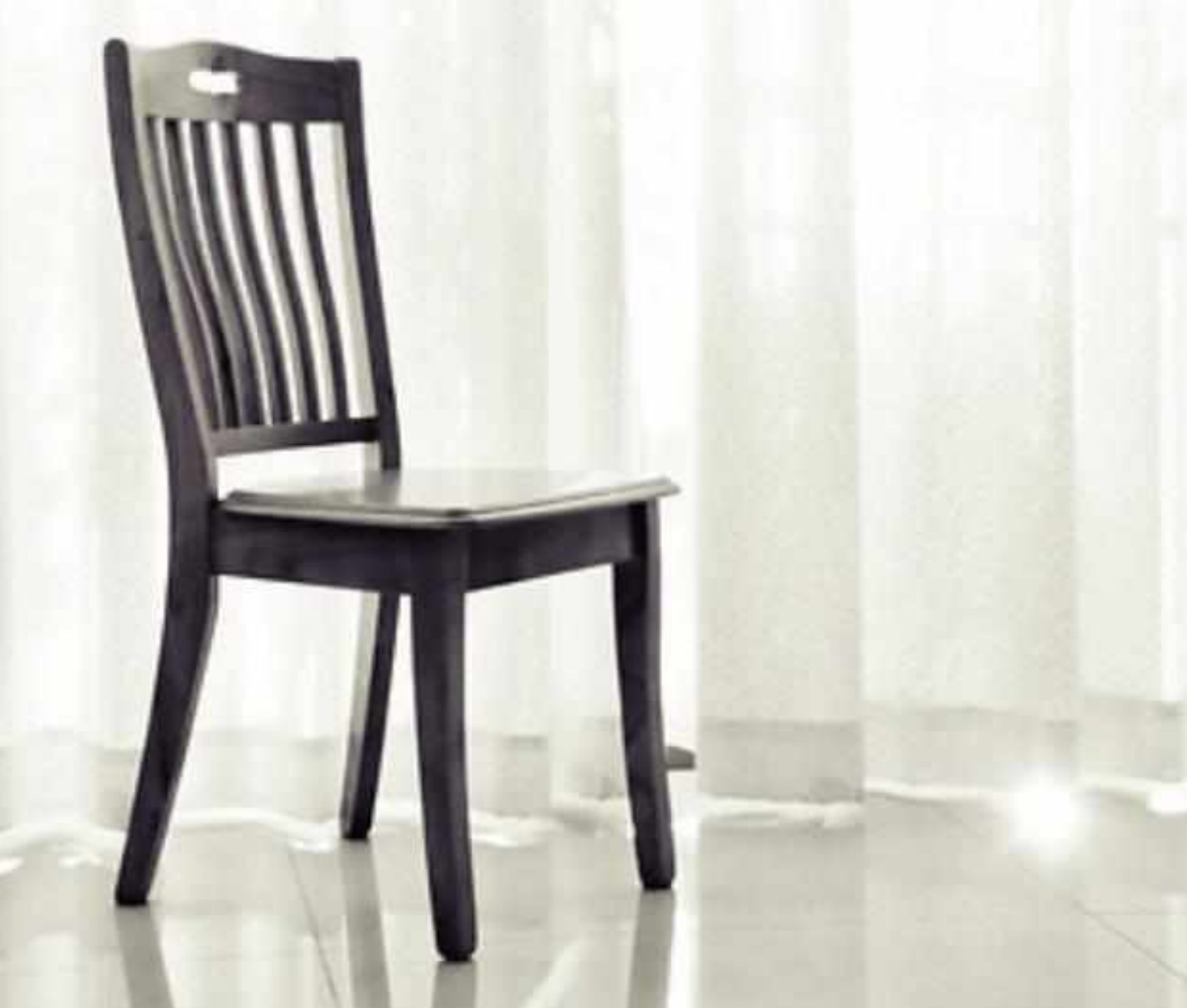}} & 
\textbf{User:} \texttt{<image>\textbackslash n}Can you elaborate on the elements of the picture provided? \\
&& \textbf{Assistant:} A wooden chair in the living room. \\ \\ \\ \\ \\ \\ \\ \\ \\ \hdashline \\

\multirow{9}{*}{SQA} & 
\multirow{9}{*}{\includegraphics[width=0.3\textwidth]{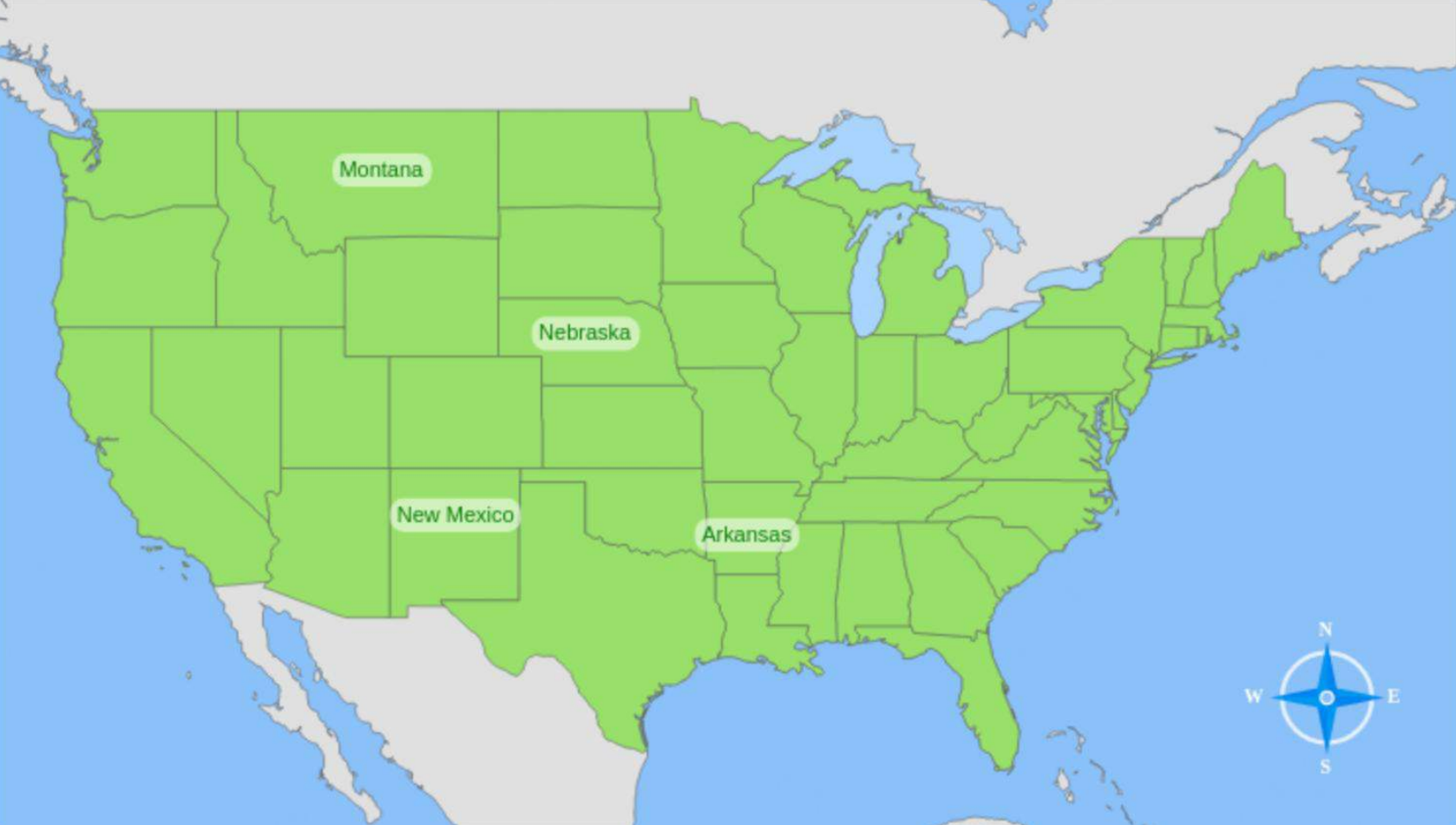}} & 
\textbf{User:}\texttt{<image>\textbackslash n}Which of these states is farthest east? 
 \texttt{\textbackslash n}Context:N/A\texttt{\textbackslash n}Options:(A) Montana (B) New Mexico (C) Arkansas (D) Nebraska\texttt{\textbackslash n} \\
&& \textbf{Assistant:} Maps have four cardinal directions, or main directions. Those directions are north, south, east, and west.\texttt{\textbackslash n}A compass rose is a set of arrows that point to the cardinal directions. A compass rose usually shows only the first letter of each cardinal direction.\texttt{\textbackslash n}The north arrow points to the North Pole. On most maps, the north is at the top of the map. To find the answer, look at the compass rose. Look at which way the east arrow is pointing. Arkansas is the farthest east. The answer is C. \\ \\ \hdashline \\

\multirow{10}{*}{VSR} & 
\multirow{10}{*}{\includegraphics[width=0.3\textwidth]{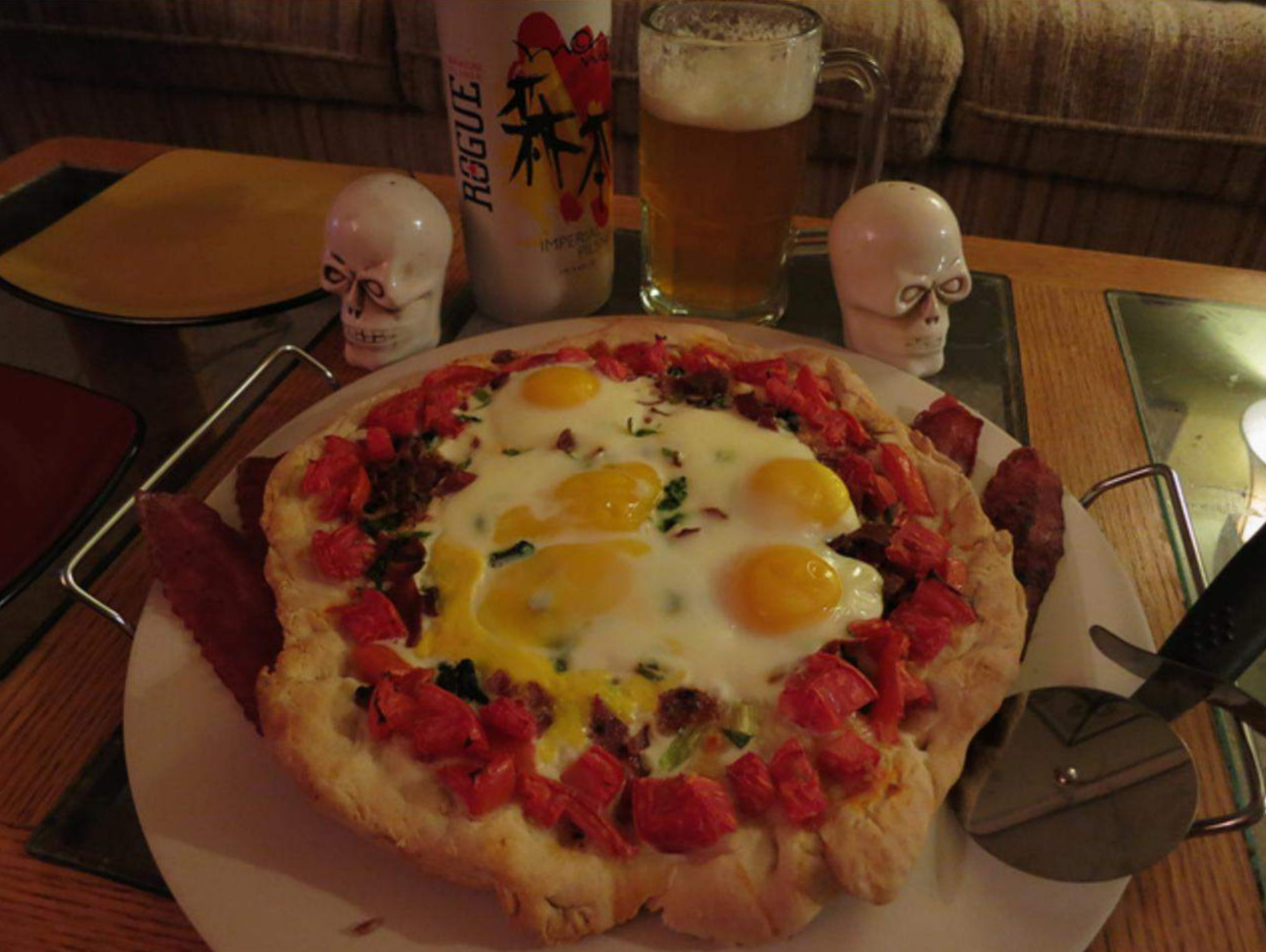}} & 
\textbf{User:} \texttt{<image>\textbackslash n}Is the pizza above the couch?\texttt{\textbackslash n}Answer the question using a single word or phrase. \\
&& \textbf{Assistant:} No. \\ \\ \\ \\ \\ \\ \\ \\ \\ \hline 

\end{tabularx}
\caption{Examples from the multi-task training data for \ours (Part 2).}
\label{tab:data formats2}
\end{table}

\clearpage 

\end{document}